\documentclass{article}

\usepackage[nonatbib,preprint]{neurips_2025}
\usepackage[dvipsnames, table, x11names]{xcolor} 
\usepackage{colortbl}

\usepackage[utf8]{inputenc} 
\usepackage[T1]{fontenc}    
\usepackage{hyperref}       
\usepackage{url}            
\usepackage{booktabs}       
\usepackage{amsfonts}       
\usepackage{nicefrac}       
\usepackage{microtype}      
\usepackage{amsmath}
 \usepackage{graphicx} 
 \usepackage{enumitem} 
\usepackage{ragged2e} 
\usepackage{tikz} 
\usepackage{tcolorbox}
\usepackage{multirow}
\usepackage{wrapfig}
\usepackage{caption}
\usepackage{makecell}

\title{HumanOmniV2: From Understanding to Omni-Modal Reasoning with Context}

\author{%
  Qize Yang$^{\dag}$\footnotemark[1] \quad Shimin Yao\footnotemark[1] \quad Weixuan Chen\footnotemark[1] \quad Shenghao Fu \quad \\ \textbf{Detao Bai\quad Jiaxing Zhao \quad Boyuan Sun\quad Bowen Yin\quad Xihan Wei\quad Jingren Zhou}\\
  Tongyi Lab, Alibaba Group\\
  \texttt{qize.yqz@alibaba-inc.com} \\
 \normalsize{\url{https://github.com/HumanMLLM/HumanOmniV2}} \\
}

\begin{document}

\maketitle
\footnotetext[1]{Equal contribution} 
\footnotetext[2]{Project lead}

\begin{abstract}
With the rapid evolution of multimodal large language models, the capacity to deeply understand and interpret human intentions has emerged as a critical capability, which demands detailed and thoughtful reasoning. In recent studies, Reinforcement Learning (RL) has demonstrated potential in enhancing the reasoning capabilities of Large Language Models (LLMs). Nonetheless, the challenges associated with adapting RL to multimodal data and formats remain largely unaddressed. In this paper, we identify two issues in existing multimodal reasoning models: insufficient global context understanding and shortcut problems. Insufficient context understanding can happen when a model misinterprets multimodal context, resulting in incorrect answers. The shortcut problem occurs when the model overlooks crucial clues in multimodal inputs, directly addressing the query without considering the multimodal information. To tackle these issues, we emphasize the necessity for the model to reason with a clear understanding of the global context within multimodal inputs. This global context understanding can effectively prevent the model from overlooking key multimodal cues and ensure a thorough reasoning process. To ensure the accurate interpretation of multimodal context information, we implement a context reward judged by a large language model, alongside format and accuracy rewards. Additionally, to improve complex reasoning capability, we employ the LLM to assess the logical reward, determining whether the reasoning process successfully integrates multimodal information with logical methods. Moreover, we develop a reasoning training dataset that incorporates context information across tasks involving images, videos, and audio. We also introduce a reasoning omni-modal benchmark, IntentBench, aimed at evaluating models in understanding complex human intentions and emotions. Our proposed method demonstrates advanced performance across multiple omni-modal benchmarks compared to other open-source omni-modal models.

\end{abstract}

\section{Introduction}
As the applications of Multimodal Large Language Models (MLLMs) in human interaction rapidly increase, understanding and reasoning human intentions and thoughts in complex scenarios becomes increasingly important.
Through recent improvements in pretraining and instruction fine-tuning, the capabilities of omni-modal models~\cite{xu2025qwen2,li2024ocean,liu2025ola,fu2025vita} have significantly advanced. 
While these models are able to handle multiple types of inputs like text, video, and audio simultaneously in complex real-world environments, they often lack strong reasoning abilities.
Inspired by DeepSeek-R1~\cite{deepseek}, many reasoning methods~\cite{eureka, video-r1,rft,vlm} for MLLMs adapt Group Relative Policy Optimization(GRPO)~\cite{grpo} to train models. Specifically, given a multimodal input and a question, these methods prompt the MLLM to generate a reasoning chain that leads to an answer. The model is optimized using both accuracy reward and format reward. The accuracy reward assesses the correctness of the answer, while the format reward encourages following the reason-answer output format.
By considering the question step-by-step, models achieve enhanced performance on various tasks, especially for multi-modal mathematical problems. However, models directly adapting vanilla GRPO rely heavily on text reasoning, ignoring the abundant multimodal cues and their comprehensive understanding.

\begin{figure}[t!]
    \centering
    \includegraphics[width=0.9\linewidth]{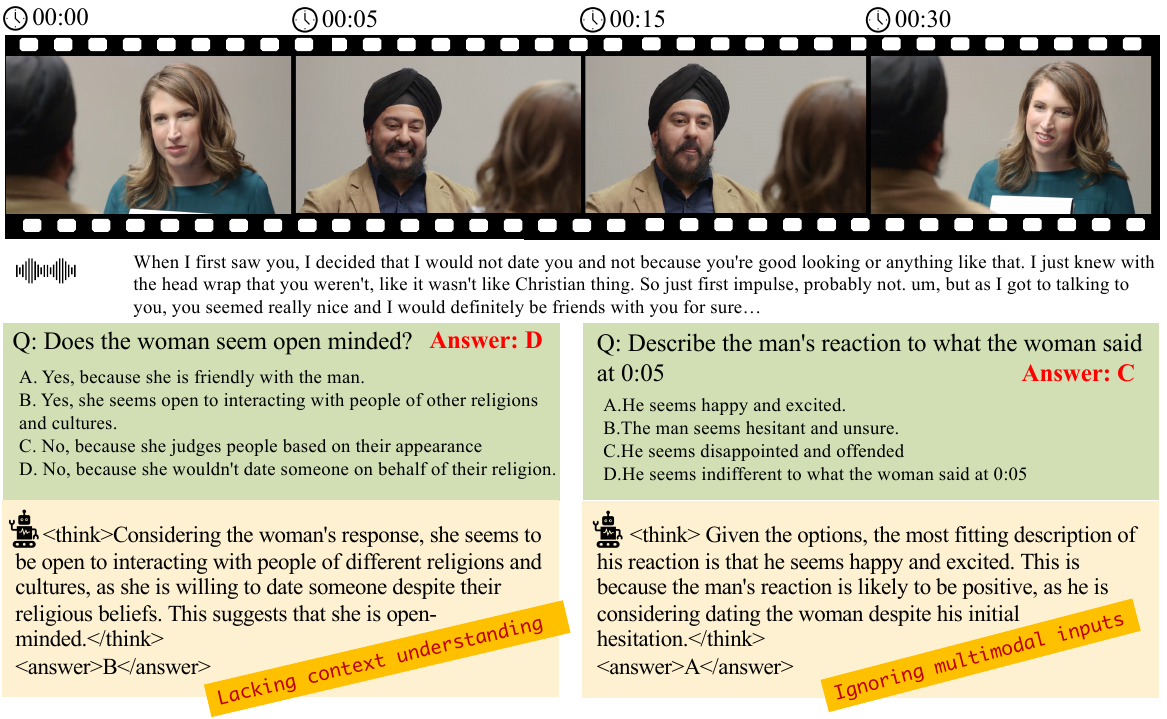}
    \caption{Visualizations of the vanilla GRPO method applied in multimodal tasks. When the model is overconfident on questions, it tends to answer questions directly without considering the global context (left) or may overlook key multimodal inputs (right).}
    \label{fig:problem}
\end{figure}

In this paper, we highlight two issues prevalent in current multimodal reasoning models: insufficient global context understanding and shortcut problems during reasoning.
Without understanding the global context or ignoring some crucial details, models will misinterpret the multimodal inputs, resulting in incorrect answers.
For example, in the left part of Figure~\ref{fig:problem}, the model trained with vanilla GRPO only captures part of the evidence in the video and provides a sub-optimal answer, failing to consider the entire video thoroughly. While in the right part, the model only observes the visual reaction of the face without considering the speech from the woman, leading to an incorrect answer.

To address these two problems, we require the model to reason based on a precise understanding of the global context within multimodal inputs. To achieve this, the model first outputs the context information within the \textit{\texttt{<context>}} tag. The context understanding prevents the model from bypassing crucial multimodal inputs and offers a comprehensive global context during the subsequent reasoning process. For example, when someone says ``no'', only within a full context can the model determine whether it is a rejection, a joke, or a reverse request. 
To ensure the model accurately comprehends the multimodal context information, we introduce a context reward in addition to the format reward and accuracy reward. 
This context reward guides the model to improve its understanding of context, assessed by a LLM that compares the consistency between the reference context and the model's output.
Additionally, to encourage the model to develop complex reasoning abilities, we introduce logical reward by using the LLM to assess whether the reasoning process integrates multimodal information and incorporates logical analysis techniques such as reflection, deduction, and induction. 
The context reward is calculated based on the context part of the completions, and the logical reward is related to both the context and reasoning parts in the completions.

Furthermore, training MLLMs to reason is extremely challenging, primarily due to the scarcity of large-scale human-annotated reasoning data \cite{step,deep,humanfb}. For this purpose, we develop a omni-modal reasoning training dataset, which incorporates context information and consists of understanding tasks involving images, videos, and audios. 
Another challenge in developing omni-modal reasoning models is the lack of related benchmarks to evaluate their performance effectively. 
We present IntentBench, a novel omni-modal benchmark designed to comprehend human activities and intentions in complex scenes. It includes 633 videos and 2,689 questions that are related to auditory and visual cues within the videos. This benchmark requires a strong understanding and reasoning of global context, careful observation, and complex social relationships. Daily-Omni~\cite{zhou2025daily} and WorldSense~\cite{hong2025worldsense} primarily focus on general perception scenarios. In these datasets, some questions are only related to either video or audio clues. In contrast, IntentBench is designed to evaluate the understanding and reasoning abilities of omni-modal models regarding the complex intentions and emotions of humans. 
Finally, we develop HumanOmniV2, which achieves the best performance among open-source omni-modal models, with scores of 58.47\% on Daily-Omni, 47.1\% on WorldSense, and 69.33\% on IntentBench.
Our contributions can be summarized as:
\begin{itemize}
    \item We propose that models should summarize the context of multimodal inputs before engaging in the reasoning process. This approach aims to mitigate issues such as skipping crucial multimodal information and context understanding on multimodal inputs. Additionally, we employ context rewards and logical rewards to incentivize models to accurately summarize the context and facilitate complex reasoning.

    \item We provide an omni-modal reasoning training dataset, which includes summaries of multimodal input and reasoning paths for both cold start training and the reinforcement learning stage. Furthermore, we have curated a human-centric benchmark, IntentBench, for omni-modal evaluation, which requires simultaneously understanding video and audio, the global context, complex social relationships, and careful observation.
    \item Our proposed HumanOmniV2 achieves the best performance across multiple omni-modal benchmarks compared to existing open-source omni-modal methods, including the newly introduced IntentBench, Daily-Omni, and WorldSense.
\end{itemize}

\section{Related Works}

\paragraph{Omni-Modal Large Language Model and Benchmarks.}
Omni-modal large language models~\cite{xu2025qwen2,zhao2025humanomni,minicpm} usually include video and audio modalities, moving towards a more comprehensive multimodal understanding. MiniCPM-o 2.6~\cite{minicpm} and Ocean-Omni-1.5~\cite{li2024ocean} enhance their vision-language foundation by adding audio processing capabilities, enabling operation across more modalities. Ola~\cite{liu2025ola} excels with its progressive modality alignment strategy, which incrementally enhances the language model's supporting modalities. VITA1.5~\cite{fu2025vita} focuses primarily on real-time, end-to-end speech interaction, while IXC2.5-OL~\cite{zhang2024internlm} introduces memory mechanisms to manage long contexts in streaming videos. ViSpeak~\cite{fu2025vispeak} proposes a novel task named visual instruction feedback in which models should be aware of visual contents and learn to extract instructions from them.

While most existing multimodal benchmarks~\cite{hu2025video,fu2025video,peng2025actionart,yue2024mmmu} focus solely on images or videos, only a few validate both audio and video simultaneously.
Although OmniBench\cite{li2024omnibench} combines image and audio for evaluation, it primarily focuses on straightforward cognitive tasks and only provides images instead of original videos. 
Daily-Omni~\cite{zhou2025daily} and WorldSense~\cite{hong2025worldsense} primarily focus on general scenarios where some challenges relate solely to video or audio clues. In contrast, our benchmark requires both audio and video understanding simultaneously to answer each question. Besides, unlike existing reasoning benchmarks that focus on the STEM (science, technology, engineering, and math) domain, our benchmark is designed to help models understand the complex intentions and emotions of humans.

\paragraph{Reinforcement Learning for MultiModal Reasoning.}

Reinforcement learning has shown greater effectiveness than supervised fine-tuning (SFT) in building general reasoning abilities. It allows models to explore a wider range of language and develop their own thinking processes. 
Recent studies~\cite{vision-r1,r1-vl} combine RL with vision and language, enhancing these capacities. Insight-V~\cite{insight} employs a multi-agent system to refine models by selecting and learning from self-generated reasoning paths.
R1-VL~\cite{r1-vl} introduces two novel rule-based reasoning rewards to improve reasoning accuracy and validity. 
Vision-R1~\cite{vision-r1} proposes a progressive thinking suppression training strategy, employing GRPO with a hard formatting result reward function to gradually enhance the model's reasoning skills on a 10k multimodal math dataset.
Video-R1~\cite{video-r1} introduces the T-GRPO algorithm, aimed at enhancing models' ability to leverage temporal information within videos for improved reasoning.
Two concurrent works, Visionary-R1~\cite{xia2025visionary} and Observe-R1~\cite{guo2025observe}, also involve the model observing the image or video first before reasoning. 
However, they focus exclusively on vision-related tasks and do not evaluate the context comprehensively, largely overlooking the potential of more comprehensive omni-modal integration. 
R1-Omni~\cite{zhao2025r1} primarily focuses on audio-visual referring segmentation, while EchoInk-R1~\cite{xing2025echoink} explores applying vanilla GRPO on OmniBench, which limits their general applicability. 
In contrast, with accurate context understanding, our omni-modal reasoning model not only understands complex human intentions but also performs exceptionally well on general omni-modal benchmarks.

\section{IntentBench}

\begin{figure}[tb!]
    \centering
    \includegraphics[width=0.9\linewidth]{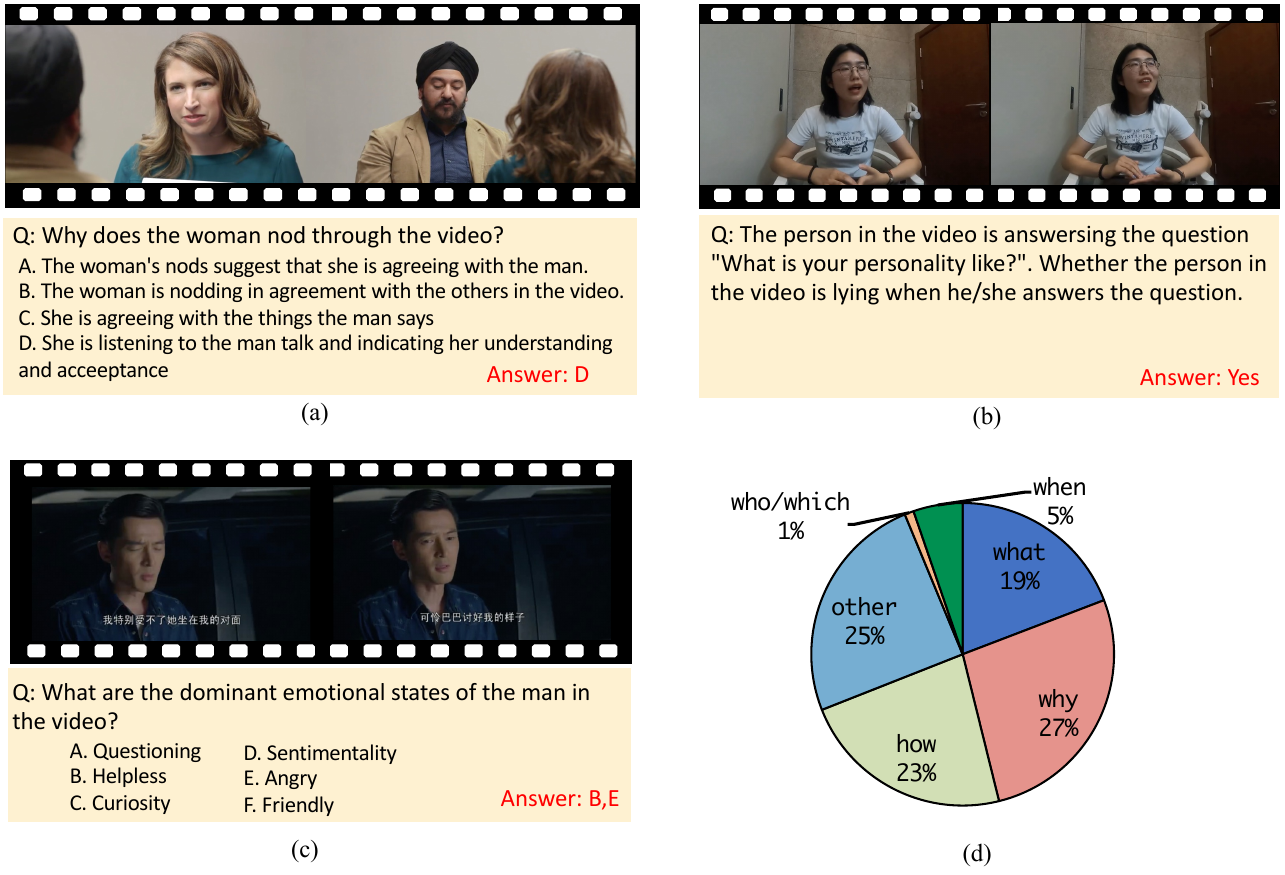}
    \caption{(a)(b)(c) are examples from Social-IQ 2.0, MDPE, and EMER, respectively. (d) is the statistic of the curated testing set from Social-IQ 2.0}
    \label{fig:dataset}
\end{figure}

Currently, multimodal reasoning evaluation datasets like MathVista~\cite{lu2023mathvista}, MMVU~\cite{zhao2025mmvu}, and VideoMMMU~\cite{hu2025video} primarily focus on the STEM domain, where audio cues are typically unnecessary. We introduce a new benchmark for evaluating omni-modal reasoning, namely IntentBench, which requires analysis on audio and vision clues. 
Our dataset focuses on understanding human social interactions in videos, including intention, emotion, and deception.
In multi-turn-taking conversations and real-world interactions, interpreting a glance, a change in tone, or the varied meanings of the same words in different contexts is a highly challenging task for multimodal large language models.
Our benchmark is curating from Social-IQ 2.0~\cite{zadeh2019social,siq2}, EMER~\cite{lian2023explainable}, and MDPE~\cite{cai2024mdpe}. 
\begin{itemize}
    \item \textbf{Social-IQ 2.0} paves the way for explainable social intelligence. The dataset is meticulously curated, featuring validated videos, questions, and answers, alongside complexity annotations for each question and answer. Social-IQ 2.0 includes more than 1,000 videos, 6,000 questions, and 24,000 answers. While humans can understand social contexts with high accuracy, current advanced computational models still face difficulties with this task. 
    \item \textbf{EMER} offers detailed explanations for its emotion, unlike traditional emotion recognition. It extracts more reliable open vocabulary emotion labels, as each label is grounded in a specific basis. It includes 332 video samples from MER2023 for annotation. Human emotions are often subtle or mixed, requiring careful observation of facial expressions, body movements, and speech. This complexity requires a deep analysis to interpret emotional states.
    \item \textbf{MDPE} includes recordings of 193 individuals, with each person answering 24 questions. Out of these, 9 questions are randomly selected for lying, and the interviewer assesses whether the candidate is lying. 
    After the interview, participants also complete the "Subject Lie Confidence Scale," where they rate their confidence in successfully deceiving, ranging from 1 to 5, with 1 meaning successful deception and 5 meaning not deceived successfully. To effectively judge deception, the model must thoroughly analyze body language, micro-expressions, tone, and speech content, which makes it a complex task.
\end{itemize}

For Social-IQ 2.0, we select 300 videos and 2,356 questions. We use GPT-4o~\cite{gpt4o} with only text modal for testing to identify challenging questions. We also replace easy options to increase the difficulty of the testing set. Finally, we conduct manual verification to ensure each question is relevant to multimodal information and cannot be answered directly through text alone. GPT-4o achieves 75.71\% on the original testing set. After our modifications, the performance is 60.02\%.
The detailed distribution of the questions of this part of the benchmark is shown in Figure~\ref{fig:dataset} (d).

For EMER, we refine the emotion vocabulary within these videos and organize all descriptive vocabulary into hierarchical categories. For the open vocabulary emotion options of the person in the videos, in addition to the original emotion ground-truth labels, we randomly select emotion description terms from other groups to create a multiple-choice question with multiple answers. We randomly select 133 videos and their corresponding questions as the testing set. We assess the performance of this part of data using the F1-score. 

For MDPE,  we reformat it to QA format and select samples where the interviewees feel uncertain about successfully deceiving—a total of 60 clips (i.e., the confidence rating is above 3). We also include 20 clips where they feel confident in their deception (i.e., the confidence rating is lower than 3, more challenging) and 120 no-deception clips, totaling 200 videos to form the deception data. 

Finally, our IntentBench comprises 2,689 questions and 633 videos. Figure~\ref{fig:dataset} shows some examples of IntentBench. Each question requires an analysis of both visual and audio cues from the videos. In the appendix, we also provide a more detailed comparison with the original dataset.

\section{From Understanding to Reasoning}



\subsection{Preliminary}
Group Relative Policy Optimization (GRPO)~\cite{grpo} streamlines the reinforcement learning approach by eliminating the critic model. This is achieved by generating multiple responses for each sample and then calculating a normalized reward within the group to determine the advantage value.
We follow GRPO with standard practice, but introduce two key modifications based on recent work~\cite{yu2025dapo,liu2025understanding}. First, we use the token-level loss to overcome the imbalance problem for long sample training. Second, we remove the question-level normalization term, which may result in varying weights in the objective for different questions, leading to a difficulty bias in optimization.
Besides, we apply the dynamic KL to avoid restricting exploration in the initial stage and diverging in the late stage, encouraging better exploration and improving training stability.
With these changes, the GRPO objective is updated as follows:
\begin{equation}
\begin{aligned}
\mathcal{J}(\theta) &= \mathbb{E}_{q \sim \mathcal{P}(q), \{o_i\}_{i=1}^G \sim \pi_{\theta_{\text{old}}}(\cdot|q)} \\ 
&\left[ \frac{1}{\sum_{i=1}^G |o_i|} \sum_{i=1}^G \sum_{t=1}^{|o_i|} \min \left( r_{i,t}(\theta) \hat{A}_{i,t}, \text{clip}(r_{i,t}(\theta), 1 - \varepsilon, 1 + \varepsilon)\hat{A}_{i,t} \right) - \hat{\beta} \text{D}_{\text{KL}} \left( \pi_{\theta} \| \pi_{\text{ref}} \right) \right] 
\end{aligned}
\end{equation}
where
\begin{equation}
\begin{aligned}
\quad \quad & r_{i,t}(\theta) = \frac{\pi_\theta(o_{i,t} \mid q, o_{i,<t})}{\pi_{\theta_{\text{old}}}(o_{i,t} \mid q, o_{i,<t})}, \quad \hat{A}_{i,t} = R_i - \text{mean}(\{R_i\}_{i=1}^G),
\end{aligned}
\end{equation}
\begin{equation}
\mathbb{D}_{K L}\left[\pi_{\theta}| | \pi_{ref}\right]=\frac{\pi_{ref}\left(o_{i} \mid q\right)}{\pi_{\theta}\left(o_{i}\mid q\right)}-\log \frac{\pi_{r e f}\left(o_{i}\mid q\right)}{\pi_{\theta}\left(o_{i}\mid q\right)}-1.
\label{eq:kl}
\end{equation}
where both $\varepsilon$ and $\hat{\beta}$ are hyper-parameters. $\varepsilon$ controls the clipping bound and limits the range of policy updates to avoid large changes that could destabilize training. $\hat{\beta}$ is the KL penalty coefficient that regularizes deviation from a reference policy $\pi_{\text{ref}}$. In this work, we dynamically reduce the KL penalty in the first $S$ iterations.
\begin{equation}
\begin{aligned}
\hat{\beta} = 
\begin{cases}
    \beta_1 + \frac{k}{S} \cdot (\beta_2 - \beta_1) & \text{if } 0 \le k \le S \\
    \beta_2 & \text{if } k > S .
\end{cases}
\end{aligned}
\end{equation}
Using a large constraint $\beta_1$ in the initial stage keeps the model within a close range of the baseline model, ensuring stable training. Conversely, using a small constraint $\beta_2$ encourages more in-depth thinking and the generation of long, detailed reasoning, but may also lead to reward hacking.

\begin{figure}[tb!]
    \centering
    \includegraphics[width=0.9\linewidth]{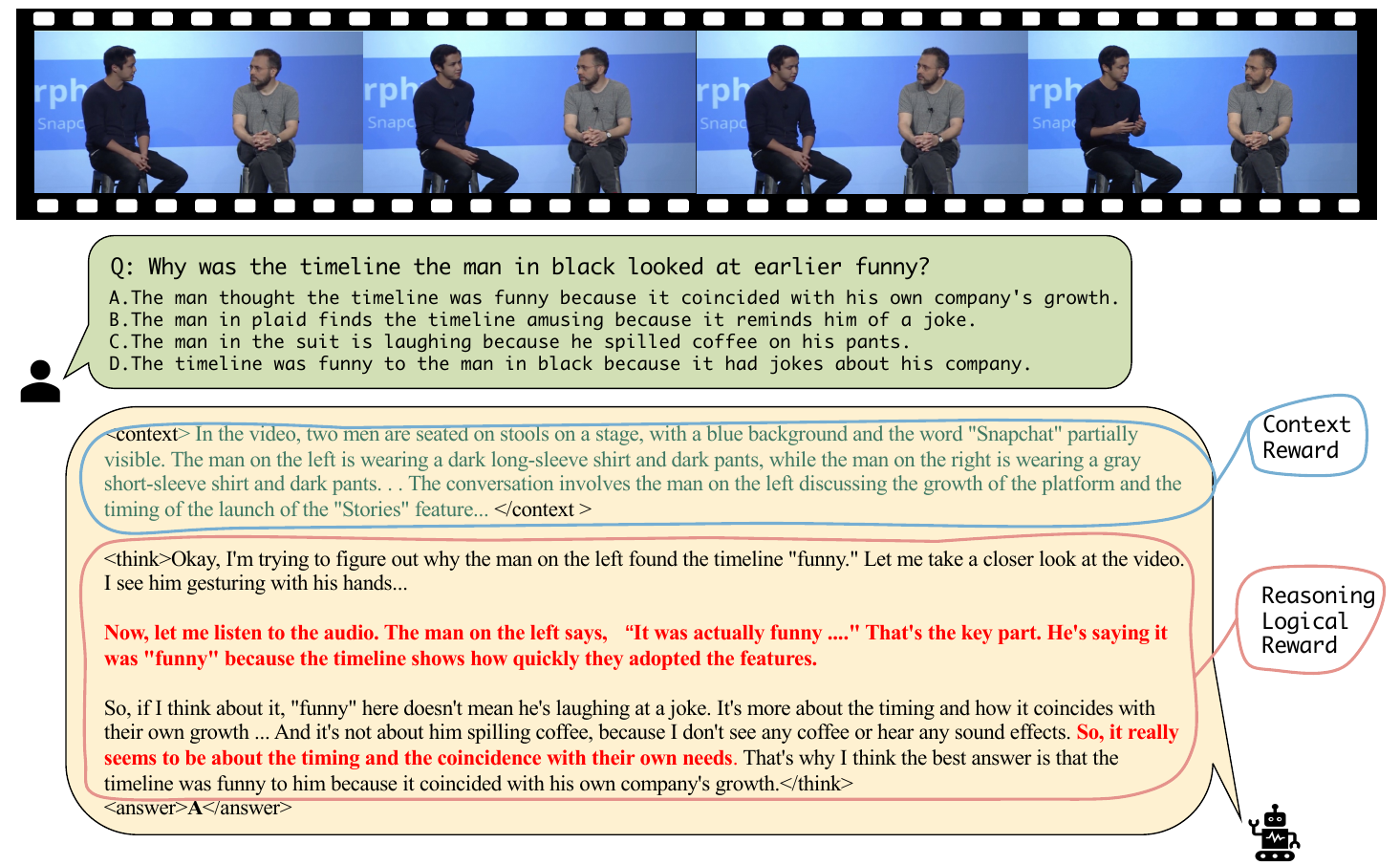}
    \caption{The reasoning path of our model on an example from Social-IQ 2.0. The model first clearly understands the context information of the video clip in the multi-person talking scenario; then it starts reasoning with the multimodal clues to precisely answer the question.}
    \label{fig:cot}
\end{figure}

\subsection{Shortcuts and Context in Multimodal Reasoning} 



When training with vanilla GRPO, one key issue is the lack of global context understanding, leading to inaccurate identification of key evidence and context during reasoning, as shown in Figure~\ref{fig:problem}. 
Another problem is that the model tends to overlook multimodal inputs, relying more on textual patterns from questions to generate answers. However, vision and audio cues are vital in multimodal interactions, and the correct answer often depends on subtle multimodal information. 

To address this challenge, we propose two modifications. First, we require the model to explicitly interpret or summarize the context information from multimodal inputs before it starts reasoning, avoiding the shortcut problem. Second, to ensure more reliable reasoning based on multimodal context, we require the model to integrate multimodal information throughout the reasoning process. We encourage reasoning capabilities such as reflection, reviewing multimodal inputs, and logical thinking.

\paragraph{Response format.}
We train the model to first generate its understanding of the multimodal context before proceeding with reasoning. Specifically, we prompt the model to generate a detailed description, which is wrapped using a \textit{\texttt{<context>...</context>}} tag, capturing the context information from multimodal inputs. In the \textit{\texttt{<think>...</think>}} tag, the model starts logically reasoning and is encouraged to reflect during reasoning, reviewing the multimodal content, as shown in Figure~\ref{fig:cot}. The model puts the final answer to the question in \textit{\texttt{<answer>...</answer>}} tag.
The final format we request the model to follow is:
\begin{center}
\begin{tcolorbox}[colback=gray!20, colframe=black!70, arc=2mm, auto outer arc, boxrule=0.5mm, width=0.9\textwidth, title=]
    \raggedright 
    \textit{\texttt{<context>...</context>}}\textit{\texttt{<think>...</think>}}\textit{\texttt{<answer>...</answer>}}\\ 

\end{tcolorbox}
\end{center}

The output is evaluated using a binary format reward  $r_f\in\{0,1\}$, which verifies whether the generated response follows the context-think-answer format. The system prompt is specified in the appendix. 
To improve the context understanding and the reasoning process, we design specific rewards for both the context understanding and reasoning phases.

\begin{figure}[tb!]
    \centering
    \includegraphics[width=0.99\linewidth]{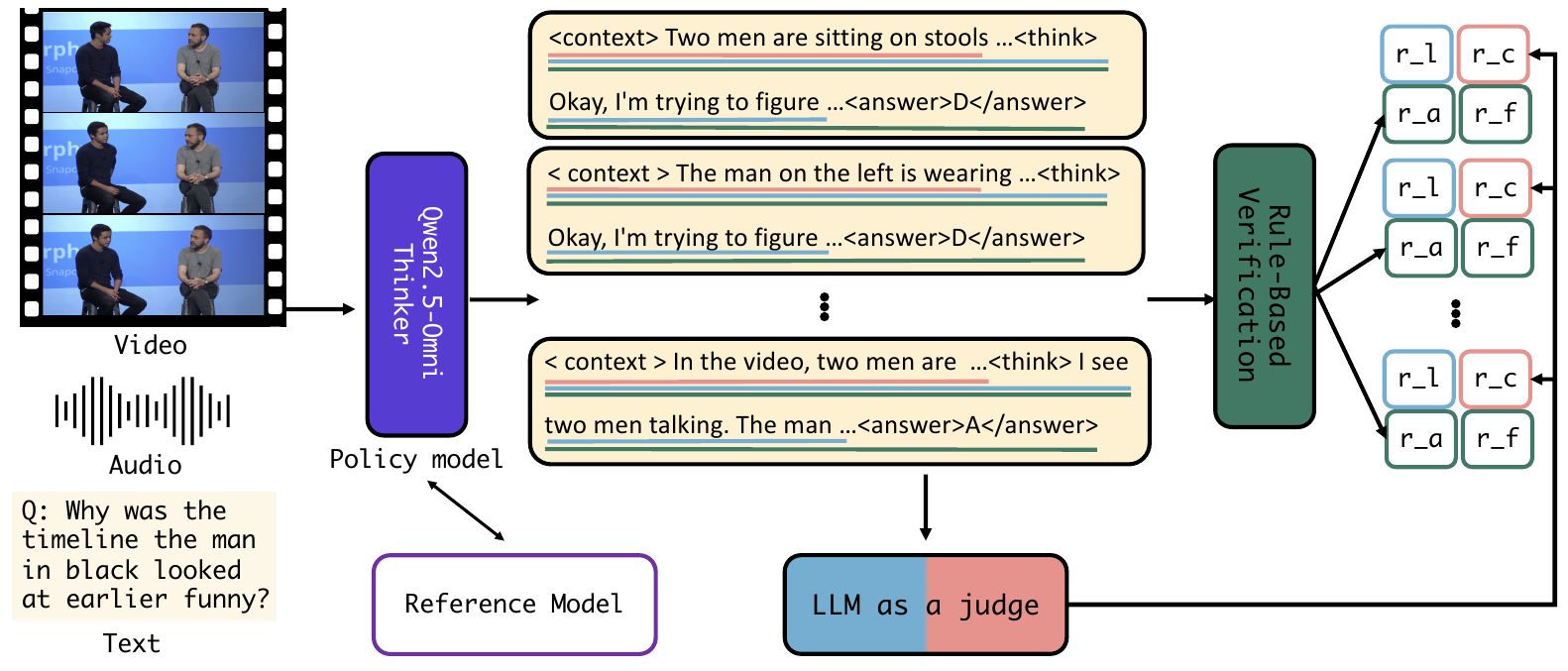}
    \caption{Illustration of our method. We use Qwen2.5-Omni-Thinker\cite{xu2025qwen2} as our base model. For each training sample, we generate 8 completions and compute format and accuracy rewards with verifiable labels. Additionally, we assess reasoning-logical and context rewards by using a LLM as the judge, applying these rewards only to corresponding seen tokens for different rewards.}
    \label{fig:main}
\end{figure}

\subsection{Context Reward and Reasoning Logical Reward with Casual Mask}
While the format enforces structure, it does not guarantee that the context is sufficiently detailed and precise to support reasoning. To address this issue, we introduce a specialized context reward $r_c\in \{0,1\}$ based on utilizing LLMs to evaluate the summary of context by offering clear scoring guidelines and prompts.
Specifically, we feed the generated context into an LLM and ask it to answer the question based on the generated context and reference context for each labeled sample. 
Furthermore, to assess the logical reward $r_l\in \{0,1\}$ of the reasoning process, we prompt the LLM to evaluate whether the reasoning incorporates multimodal information, reflection, confirmation, and logical deduction. 
Since these two rewards only evaluate certain intermediate tokens within the reasoning path, the context reward and logical reward are masked and applied solely to the corresponding seen tokens, as shown in Figure~\ref{fig:main}. To ensure balanced training, the context reward and logical reward are individually normalized.

For accuracy rewards, we utilize LLM to obtain similarity scores for open-ended question answers, rather than relying on metrics such as BLEU~\cite{papineni2002bleu} or ROUGE~\cite{lin2004rouge} scores. For multiple-choice questions with a single answer, we use accuracy as the reward. For multiple-choice questions with multiple answers, we utilize the F1-score as the reward. For OCR and ASR tasks, the reward is calculated as $1-\text{WER (Word Error Rate)}$. In evaluating numerical answers, we check for equality between two numbers.



\subsection{Training Recipe}

\noindent\textbf{Cold start training.} We select Qwen2.5-Omni-7B-thinker as our base model and modify the model's system prompt to ensure it follows our specified output format. During the initial stage, we use long CoT (Chain-of-Thought) data, incorporating video and image reasoning data for cold start training, to stabilize the model in the reasoning phase. 


\noindent\textbf{RL Training Stage 1.} After completing the cold start training, we sample 8 times for each sample and retain those with an accuracy within $(0, 0.75)$. 
At this stage, all RL training samples are annotated with context to facilitate rewarding relevant content during RL training. Thus, the reward would be more dense.
For reasoning rewards, we use the context generated in each completion instead of the ground-truth context. During this phase, we employ LLM (\textit{e.g.}, Qwen2.5-72B) as the reward model. For detailed prompt information, please refer to the supplementary material.

\noindent\textbf{RL Training Stage 2.}
After completing the first stage of RL training, we aim to enhance the model's general capabilities by utilizing general RL training data, including the data sourced from Video-R1\cite{feng2025video} and OmniInstruct~\cite{li2024omnibench}. For these datasets, we also sample 8 times for each sample and retain those with an accuracy within the range of $(0, 0.75)$. Since the previous phase already guides the model to learn more precise context and complex reasoning processes, in this phase, we only use the format reward and accuracy reward.

\vspace{-0.3cm}
\section{Experiments}
\subsection{Experimental Settings}

\noindent\textbf{Training data, details, and baselines.} %
To obtain high-quality CoT data, we sample part of the data from Video-R1, the training set of Social-IQ2.0, and the remaining 200 entries from EMER. Then, we use Gemini-2.5-pro~\cite{gemini2.5pro} to rewrite the reasoning processes, and for Social-IQ2.0, we carry out manual reviews to ensure the reasoning accuracy. 
For the sampled data from Video-R1, we also use Gemini-2.5-pro to rewrite the reasoning path. Finally, we collect 24K high-quality video-audio training data for cold-start training and RL training. 
In the cold start training stage, the learning rate is set to $1e^{-5}$ and the batch size is 128. 
During the RL stages, the clip range for training is 0.2. The step $S$ for KL constraints is half of the total steps, with $\beta_1$ and $\beta_2$ set at 0.04 and 0.01, respectively. The learning rate is $1e^{-6}$, and the maximum output length is 2048.
To validate the effectiveness of our proposed method, we compare our full method against the model trained only with SFT using CoT data, our method without extra rewards, and the vanilla GRPO~\cite{grpo}. 

\noindent \textbf{Evaluation and benchmarks.}
In testing, the video frame sampling rate is 2 FPS, with a maximum of 32 frames.
We evaluate our method on IntentBench, Daily-Omni~\cite{zhou2025daily}, and WorldSense~\cite{hong2025worldsense}. Most questions in these benchmarks require simultaneous understanding of both video and audio, making them well-suited for assessing omni-modal models. 

Daily-Omni~\cite{zhou2025daily} is an audio-visual questioning and answering benchmark that includes 684 videos capturing everyday life scenarios from various sources. These videos are rich in both audio and visual information, and the benchmark features 1,197 multiple-choice QA pairs spread across 6 major tasks.

WorldSense~\cite{hong2025worldsense} features a strong coupling of audio and video, requiring models to effectively leverage the synergistic perception capabilities of omni-modal data.  It includes 1,662 audio-visual synchronized videos, sorted into 8 main domains. Additionally, there are 3,172 multiple-choice QA pairs across 26 different tasks to enable thorough evaluation.

\begin{table}[t!]
\centering
\caption{Comparison with other methods on Daily-Omni~\cite{zhou2025daily}.}
\scriptsize

\setlength{\tabcolsep}{2.5pt}
\begin{tabular}{lc|cccccc|cc|c}
\toprule
{Methods}  & \makecell{LLM \\ Size} & {{\makecell[c]{AV Event\\Alignment}}} & {{Comparative}} & {{\makecell[c]{Context\\Understanding}}} & {{\makecell[c]{Event\\Sequence}}} & {{Inference}} & {{Reasoning}} & {{\makecell[c]{30s\\Subset}}} & {{\makecell[c]{60s\\Subset}}} &{{Avg}}  \\
\midrule
\multicolumn{10}{c}{\textit{\textbf{Proprietary MLLMs}}}\\
\midrule
Gemini 2.0 Flash & - & {62.18} & {73.28} & {63.73} & {63.72} & {76.62} & {75.43}  & 62.29 & 56.57 & {67.84}\\
Gemini 2.0 Flash Lite & - & {55.04} &64.89 & {58.03} & {54.25} & {74.03} &72.00  & 60.57 & 53.01 & 61.32\\
\midrule
\multicolumn{10}{c}{\textit{\textbf{Open-Source Video-Audio MLLMs}}}\\
\midrule
Unified-IO-2~\cite{lu2024unified} &8B  & 25.63 & 31.30 & 26.42 & 25.82 & 35.06 & 29.71 & 26.74 & 30.00 &28.24  \\
VideoLLaMA2 ~\cite{cheng2024videollama}& 7B & 35.71 & 35.88 &35.75 & 31.70 & 40.91 & 34.29 & 38.02 & 31.82 & 35.17\\
Qwen2.5-Omni~\cite{xu2025qwen2}& 3B & 38.66 & 48.09 & 33.68 & 33.99 & 54.55 & 44.00  & 46.68 &48.36 & 40.52\\
Qwen2.5-Omni~\cite{xu2025qwen2}& 7B & 44.12& 51.15 & 38.86 & 40.52& 57.79 & 61.71  & 42.35 & 38.36 & 47.45\\
Ola~\cite{liu2025ola} & 7B & 40.33 & 60.30 & 39.89 & 44.11 & 61.03 & 66.28 & 50.85 & 48.72 & 49.87\\
MiniCPM-o~\cite{liu2025ola} & 7B & 40.33 & 61.06 & 49.22 & 48.36 & 68.83 & 61.14 & 54.09 & 52.00& 53.13\\
\rowcolor[HTML]{EFEFEF}
\textbf{Ours} & 7B & \textbf{46.63} &  \textbf{67.93} & \textbf{51.81} & \textbf{51.63} &\textbf{72.72}&  \textbf{74.28} & \textbf{63.06}  & \textbf{53.09}& \textbf{58.47} \\

\bottomrule
\end{tabular}
\label{tab:daily} 
\end{table}

\subsection{Main Results}
We compare our method with open-source omni-modal methods and proprietary MLLMs on Daily-Omni and WorldSense, as shown in Table~\ref{tab:daily} and Table~\ref{tab:world}, and IntentBench in Table~\ref{tab:intent}. The experimental results indicate that our model outperforms most open-source omni-modal models. Specifically, in out-of-domain testing, our method achieves the best results among open-source models. Particularly, it boosts performance from 61.71 to 74.28 in the "reasoning" task within Daily-Omni, demonstrating comparable performance to Gemini-2.0-Flash-Lite. We also observe that our method performs worse than Qwen2.5-Omni in the ``Performance'' and ``Music'' tasks of WorldSense. These two tasks are primarily related to visual perception or auditory perception.
In in-domain testing, our method achieves the best results compared to other methods and baselines, as shown in Table~\ref{tab:intent}. Our method effectively combines global context information with fine-grained audio-video clues during reasoning, leading to higher accuracy in complex social situations and understanding human intent.

\begin{figure}[tb!]
    \centering
    \includegraphics[width=0.95\linewidth]{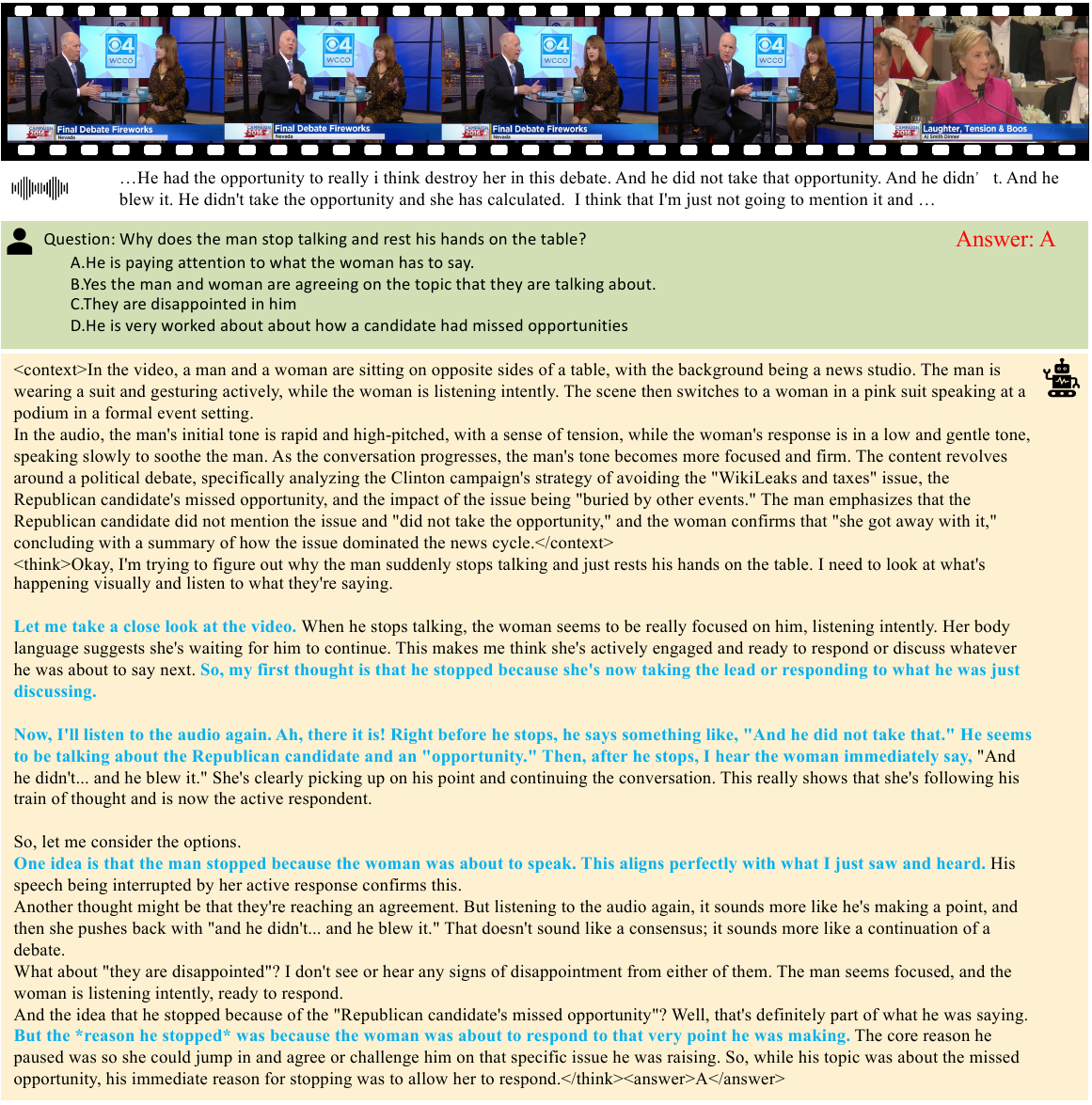}
    \caption{Visualization result of our method on IntentBench.}
    \label{fig:case1}
\end{figure}

As shown in Figure~\ref{fig:case1}, we present an example from our IntentBench. Within the \textit{\texttt{<context>}} tag, our model summarizes the global context information, including who, where, and what is happening in the video. This global context information helps mitigate the shortcut problem in multimodal reasoning and provides a basis for subsequent reasoning. Consequently, the model avoids focusing solely on specific details, ensuring that it does not overlook the broader context during reasoning.

\begin{table}[t]
\centering
\scriptsize
\caption{Comparison with other methods on WorldSense~\cite{hong2025worldsense}.}
\label{tab:world}
\begin{tabular}{lc|cccccccc|c}
\toprule

Methods & \makecell{LLM \\ Size} & \makecell{Tech \& \\ Science}  & \makecell{Culture \& \\ Politics} & \makecell{Daily \\ Life} & \makecell{Film \& \\ TV} & \makecell{Perfor-\\mance} & Games & Sports & Music & Avg \\

\midrule
\multicolumn{11}{c}{\textit{\textbf{Proprietary MLLMs}}}\\
\midrule
Claude 3.5 Sonnet~\cite{claude}  & - & 43.7 & 31.7 & 30.6 & 36.5 & 30.7 & 31.9 & 36.6 & 33.9 & 34.8  \\
GPT 4o~\cite{hurst2024gpt}  & - & 48.0 & 44.0 & 38.3 & 43.5 & 41.9 & 41.2 & 42.6 & 42.7 & 42.6 \\
Gemini 1.5 Pro~\cite{team2024gemini}  & - &  {53.7} & 47.2 &  {50.3} &  {50.4} &  {52.4} &  {46.8} & 40.2 & 42.0 &  {48.0} \\

\midrule
\multicolumn{11}{c}{\textit{\textbf{Open-Source Video-Audio MLLMs}}}\\
\midrule
Unified-IO-2 XXL~\cite{lu2024unified} & 7B & 27.1 & 31.7 & 23.9 & 23.7 & 25.5 & 23.7 & 25.7 & 27.3 & 25.9 \\
VideoLLaMA2~\cite{cheng2024videollama} & 7B & 29.4 & 25.4 & 21.8 & 24.5 & 26.2 & 24.6 & 25.5 & 27.1 & 25.4 \\
VITA-1.5~\cite{fu2025vita} & 7B & 38.2 & 35.9 & 34.3 & 39.8 & 41.2 & 32.6 & 34.7 & 39.9 & 36.9 \\
Qwen2.5-Omni~\cite{xu2025qwen2} & 7B & 47.8 &  {49.8} & 43.6 & 43.8 & \textbf{48.3} & 39.1 & {43.5} &  \textbf{47.3} & 45.4 \\
\rowcolor[HTML]{EFEFEF}
\textbf{Ours} & 7B & \textbf{50.2} &  \textbf{51.7} & \textbf{47.6} & \textbf{44.8} & 47.3 & \textbf{44.3} &  \textbf{45.2} & {44.2} & \textbf{47.1}\\
\bottomrule
\end{tabular}
\end{table}

\begin{table}[t]
\centering
\scriptsize

\caption{Comparison with other methods on our IntentBench. $^*$Since these models do not support audio input, we use the transcript of the audio instead. ``ER'' represents the context reward and logical reward.
}
\label{tab:intent}
\resizebox{0.99\textwidth}{!}{ 
\begin{tabular}{lc|cccccc|c|c|c}
\toprule
\multirow{2}{*}{Methods} & \multirow{2}{*}{LMM}  & \multicolumn{6}{c|}{Social} &  \multirow{2}{*}{Emotion} &  \multirow{2}{*}{Deception} &  \multirow{2}{*}{Avg} \\ 
 &   & Why & How & What  &When &Who/Which  &Other & & & \\ 
\midrule
\multicolumn{11}{c}{\textit{\textbf{Proprietary MLLMs}}}\\
\midrule
GPT-4o$^*$~\cite{gpt4o}        & -       &    61.46    & 55.69    &  60.00 &35.71    & 76.00   & 63.31   &   60.99   &59.00    &  59.98 \\
GPT-o1$^*$\textit{(think)}~\cite{gpto1}        & -       &  68.19      &  65.82   &   66.04    & 57.14      &  76.00  & 68.83   & 67.26     & 59.50    & 66.69  \\

Gemini-2.5-Pro \textit{(think)}~\cite{gemini2.5pro}   & -       &   68.57     &  67.41    & 65.12      & 57.14      & 64.00   & 70.03  &   68.23    &60.00    &67.15   \\
\midrule
\multicolumn{11}{c}{\textit{\textbf{Open-Source Video-Audio MLLMs}}}\\
\midrule
MiniCPM-o~\cite{minicpm}       & 8B      &  57.87      &  53.48    &   57.14    & 57.14      &  68.00  & 61.14  &    23.85   &  49.50  &  54.51 \\
VITA-1.5~\cite{fu2025vita}       & 7B      &  53.15      &49.36      &   51.66    & 71.42      &  64.00  &  61.14 &    53.20   &  59.50  &  54.17 \\
Ola~\cite{liu2025ola}       & 7B   &  60.60      &  55.37    &   56.87    &   64.28   & 76.00  &    62.91   &  46.66  & 44.50 &57.41 \\
Qwen2.5-Omni~\cite{xu2025qwen2}      & 7B    &   62.60     &   63.44   &  63.53  &    57.14   &  76.00  & 69.03  &    59.74   &63.50    &  64.20 \\
\midrule
\multicolumn{11}{c}{\textit{\textbf{Ours}}}\\
\midrule
\rowcolor[HTML]{EFEFEF}
\textbf{Ours (full method)}                       & 7B     &  66.76      &  \textbf{67.08}    &  \textbf{71.25}     &   50.00    & \textbf{84.00}   & \textbf{72.39}  & 82.41      &  \textbf{64.00}  & \textbf{69.33}  \\
\textit{- SFT (CoT)}                & 7B     &   58.30     & 60.28    &   62.91    & \textbf{64.28}      &  76.00  & 65.28  &  79.00     &  57.00  &  62.03 \\
\textit{- Ours w/o ER}                             & 7B  & \textbf{67.47}   &  65.98      &   68.54   &  50.00     &  76.00     & 71.20   & \textbf{84.00}  &   62.00    &   68.44 \\
\textit{- Ours w/o context and ER}                           & 7B      &    64.47    &  64.08   &  66.45     &   50.00    & 80.00  & 71.40  &   81.37    &   61.50 &66.72   \\

\bottomrule
\end{tabular}
}
\end{table}

\subsection{Ablation Study and Analyses}
\textbf{The effect of cold start.}
As shown in Table~\ref{tab:intent}, the performance of the cold start long CoT training (i.e.,  ``\textit{SFT(CoT)}'' row in Table~\ref{tab:intent}) slightly decreases compared to Qwen2.5-Omni. We think the primary reason is that the CoT training data is largely smaller than Qwen2.5-Omni (1,200 billion tokens in total). Additionally, Qwen2.5-Omni already performs comparably well on original Social-IQ 2.0 and MDPE, possibly because the training sets of these datasets are already included in the training of the original Qwen2.5-Omni.

\textbf{The effect of context reward and logical reward.}
The context reward incentivizes the model to enhance context quality, while the logical reward promotes complex reasoning and integrates multimodal inputs into the reasoning process. Consequently, as shown in Table~\ref{tab:intent}, the performance without these additional rewards is inferior to our full method.

\textbf{The effect of context.}
In Table~\ref{tab:intent}, while our model without context and extra rewards performs better than Qwen2.5-Omni, its performance drops to 66.72 compared to our full method. This result highlights the importance of the global understanding of multimodal inputs. The context can effectively offer a comprehensive understanding for the subsequent reasoning process.

\vspace{-0.2cm}
\section{Conclusion}
\vspace{-0.2cm}
In this paper, we highlight two issues in multimodal reasoning: insufficient global context understanding and the tendency to overlook multimodal inputs. To address these challenges, we introduce HumanOmniV2, an omni-modal reasoning model that requires summarizing the global context of inputs. By basing reasoning on this comprehensive understanding, it reduces problems like overlooking essential multimodal information and enhances context comprehension of multimodal inputs.
We also introduce an omni-modal reasoning benchmark that focuses on understanding human intentions and emotions in complex social interactions. Our proposed method surpasses other open-source omni-modal methods and vanilla GRPO across various benchmarks, demonstrating the effectiveness of summarizing the context of multimodal inputs before reasoning.

\section{Limitations}
Our method reduces the probability of shortcuts and context errors during reasoning, but it does not completely solve the issues. 
This response format might also restrict the ability of the model to correct context information during the following thinking process. 
Moreover, our experiments are primarily based on a 7B model, so consistent conclusions may not be guaranteed at larger scales. Additionally, shorter response length also restricts the model's ability to reflect thoroughly on multimodal information. 
In future research, along with expanding context and pretraining scale, we aim to explore methods for multiple verification of multimodal information during the reasoning process to enhance accuracy.

\section{Appendix}

\subsection{More visualizations of our method}
\begin{figure}[h]
    \centering
    \includegraphics[width=0.99\linewidth]{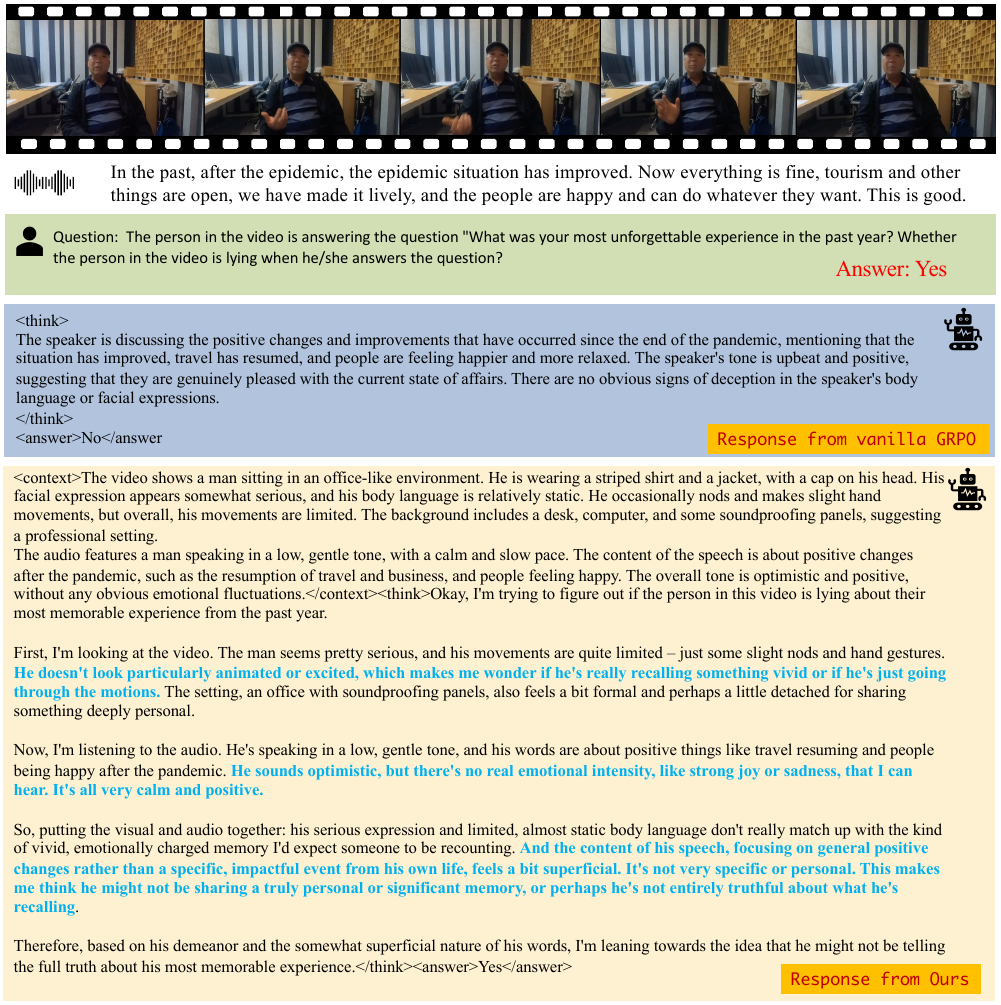}
    \caption{Visualization result of our method and the compared method on IntentBench.}
    \label{fig:case2}
\end{figure}

\begin{figure}[h]
    \centering
    \includegraphics[width=0.99\linewidth]{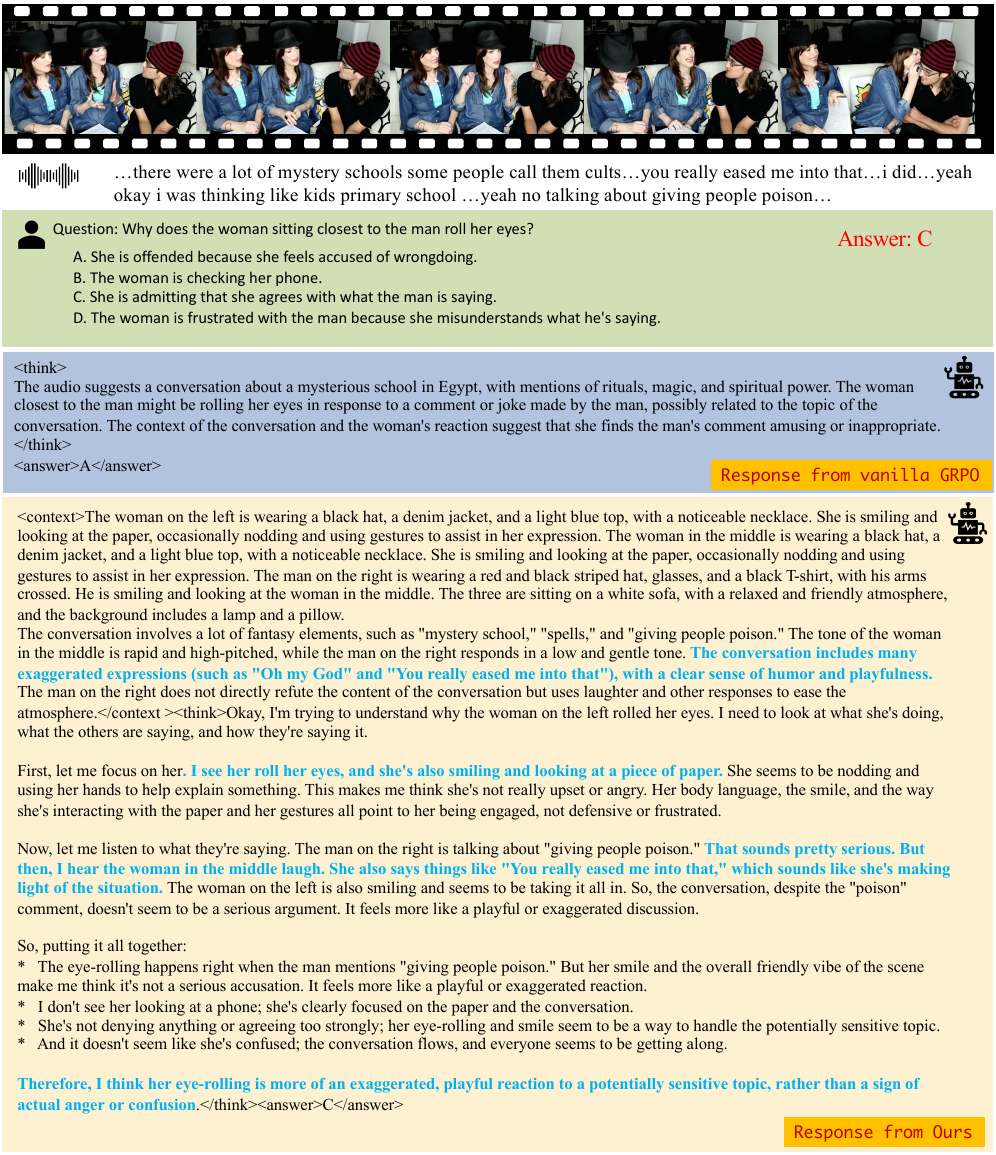}
    \caption{Visualization result of our method and the compared method on IntentBench.}
    \label{fig:case3}
\end{figure}

\begin{figure}[h]
    \centering
    \includegraphics[width=0.99\linewidth]{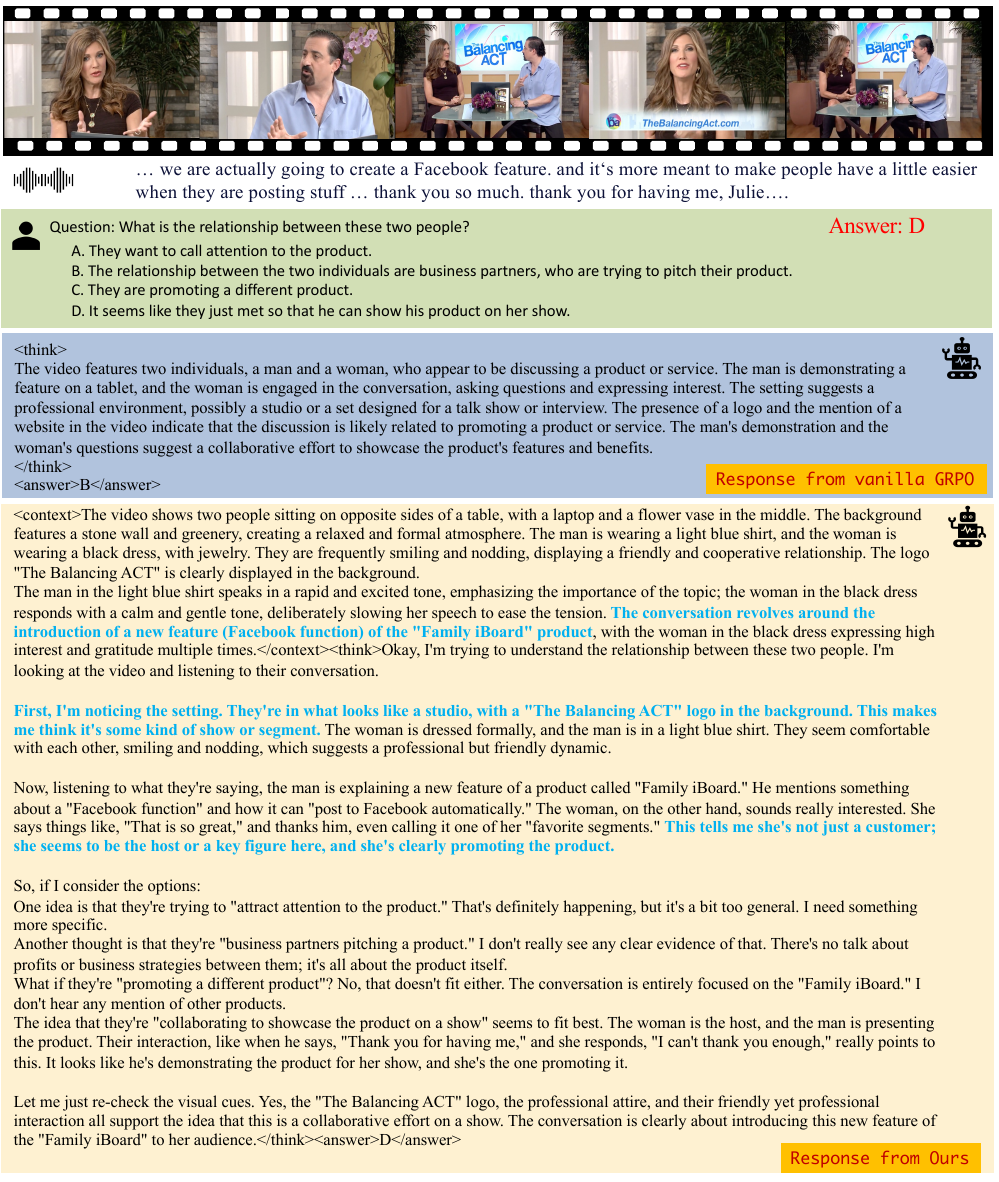}
    \caption{Visualization result of our method and the compared method on IntentBench.}
    \label{fig:case4}
\end{figure}

\begin{figure}[h]
    \centering
    \includegraphics[width=0.99\linewidth]{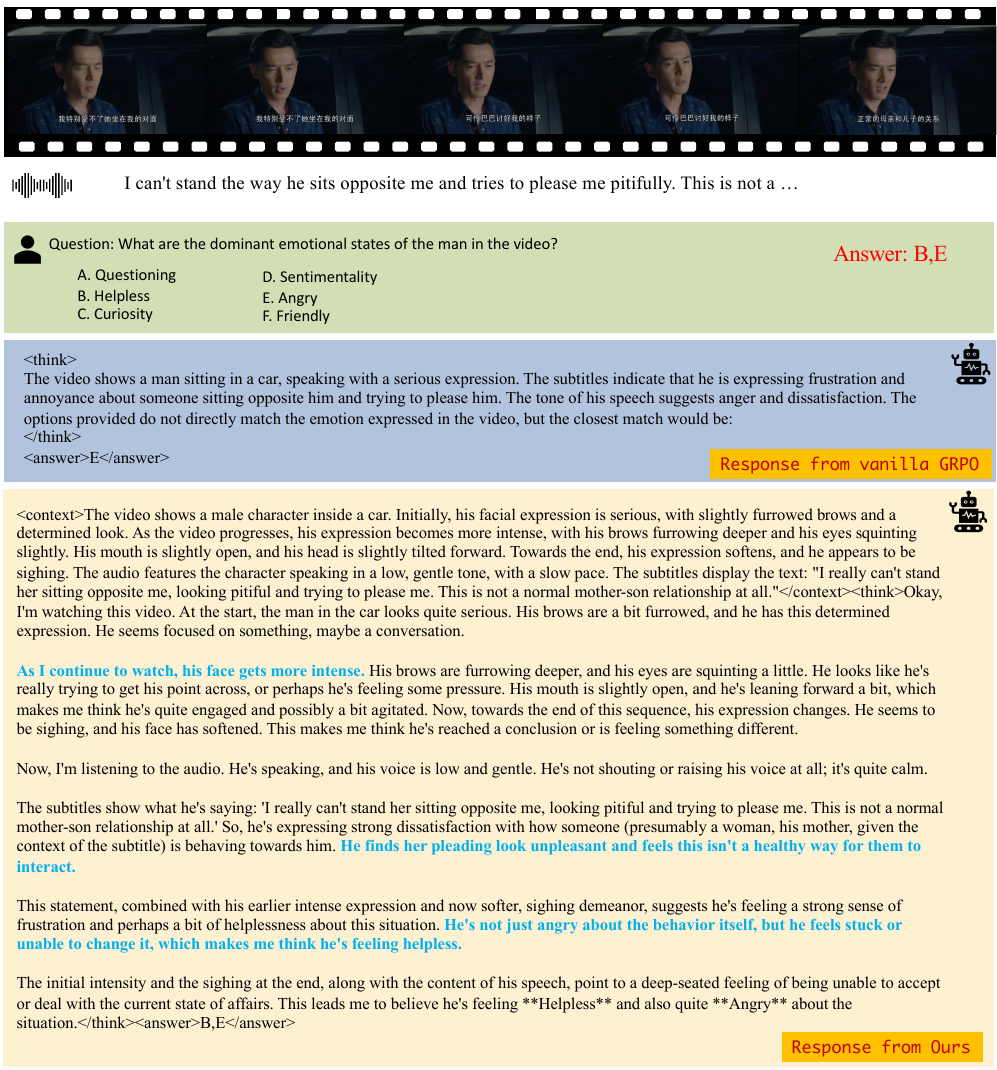}
    \caption{Visualization result of our method and the compared method on IntentBench.}
    \label{fig:case5}
\end{figure}

From Figure~\ref{fig:case2} to \ref{fig:case5}, we present visualizations of the evaluation results for our full method and the compared method on IntentBench. Our model can effectively comprehend global context information and capture fine-grained video-audio clues from multimodal inputs during the reasoning process.

\newpage
\newpage
\newpage
\newpage

\subsection{The prompts used in our method}

As shown in Figure~\ref{fig:system_prompt_box}, we carefully design the system prompt to guide our model in thoroughly understanding the global context while encouraging self-reflection and verification, ensuring that the model focuses on and adheres to our specified output format. Figure~\ref{fig:plr} and Figure~\ref{fig:pcr} are the prompts for assessing logical reward and context reward, respectively. 
We carefully design the user prompt to guide the LLM in precisely assessing the quality of the context summary and determining whether the reasoning incorporates multimodal information, reflection, confirmation, and logical deduction.

\begin{figure}[h!] 
    \centering 
    \small 

    \begin{tcolorbox}[colback=gray!20, colframe=black!70, arc=2mm, auto outer arc, boxrule=0.5mm, width=0.9\textwidth, title=]
    
    You are a helpful assistant. Your primary goal is to deeply analyze and interpret information from various available modalities (image, video, audio, text context) to answer questions with human-like depth and a clear, traceable thought process.\\

    Begin by thoroughly understanding the image, video, audio, or other available context information, and then proceed with an in-depth analysis related to the question. \\

    In reasoning, it is encouraged to incorporate self-reflection and verification into your reasoning process. You are encouraged to review the image, video, audio, or other context information to ensure the answer accuracy.\\

    Provide your understanding of the image, video, and audio between the <context> </context> tags, detail the reasoning between the <think> </think> tags, and then give your final answer between the <answer> </answer> tags. 
    
    \end{tcolorbox}
\caption{The system prompt of our model.}

\label{fig:system_prompt_box} %
\end{figure}

\begin{figure}[h!]
\centering 
\small
\begin{tcolorbox}[colback=gray!20, colframe=black!70, arc=2mm, auto outer arc, boxrule=0.5mm, width=0.9\textwidth, title=]
Please analyze whether the reasoning text is derived from the evidence and context text based on the following criteria and give a score of 0-5:
Grading criteria description (relevance and rationality):

Integration of Clues (1 point): During the reasoning process, there is incorporation of clues from the video, image, or audio.\\

Reflection and Confirmation (1 point): The reasoning involves reflection or second confirmation of choices or answers, including revisiting video, image, or audio evidence.\\

Logical Reasoning (1 point): The thought process is clear, deriving conclusions through rigorous logical reasoning, analysis, or extension without additional assumptions or contradictions.\\

Problem Analysis (1 point): The reasoning process includes thorough analysis in conjunction with the problem at hand.\\

Overall Consistency (1 point): The reasoning text is based on visual or audio evidence and context information, presenting no extra assumptions or contradictions.\\

Assign one point for each criterion that is met, for a total possible score of five points. Verify that each criterion is addressed and reflect this in your scoring.\\

context: {reference}\\
reasoning path: {hypothesis}\\

only return the score number:

\end{tcolorbox}
\caption{The prompt for assessing the logical reward.} 
\label{fig:plr} %
\end{figure}

\begin{figure}[h!] 
\centering 
\small
\begin{tcolorbox}[colback=gray!20, colframe=black!70, arc=2mm, auto outer arc, boxrule=0.5mm, width=0.9\textwidth, title=]
You are assessing how well the `hypothesis' text covers the key information from the `reference' text. Differences in wording or extra details in the `hypothesis' are fine if the reference's main points are included:\\

Score based on this coverage:\\

5 points : Hypothesis clearly and accurately reflects significant core themes or key aspects of the reference. It demonstrates a good understanding of a substantial part of the reference material.

4 points : Hypothesis reflects some important themes or aspects of the reference. The connection is evident, though perhaps not as comprehensive or central as a 5.

2 points : Hypothesis shows a recognizable connection to themes or aspects of the reference, but it might be more superficial, focus on less central points, or only partially grasp a key aspect.

1 points : Hypothesis has a tenuous or very limited connection to the reference. It might touch on a peripheral detail or a heavily reinterpreted aspect, but largely misses the main substance.

0 points : Hypothesis does not reflect any significant themes or key aspects of the reference, or is on a completely different topic.\\

Example analysis process:

Identify main themes and key aspects in `reference'.\\
Determine if `hypothesis' connects to or discusses any of these themes/aspects from `reference'.\\
Judge the strength and relevance of this connection. Is a core part of the `reference' reflected?\\
Differences are expected; evaluate if the `hypothesis' still meaningfully reflects some key part of the `reference'.\\

Assign score based on how well a significant aspect is reflected.

reference: {reference}\\
hypothesis: {hypothesis}\\

only return the score number:

\end{tcolorbox}
\caption{The prompt for assessing the context reward.} 
\label{fig:pcr} %
\end{figure}

\newpage
\subsection{The Differences between IntentBench and Its Original Datasets.}

\begin{figure}[t!]
    \centering
    \includegraphics[width=0.8\linewidth]{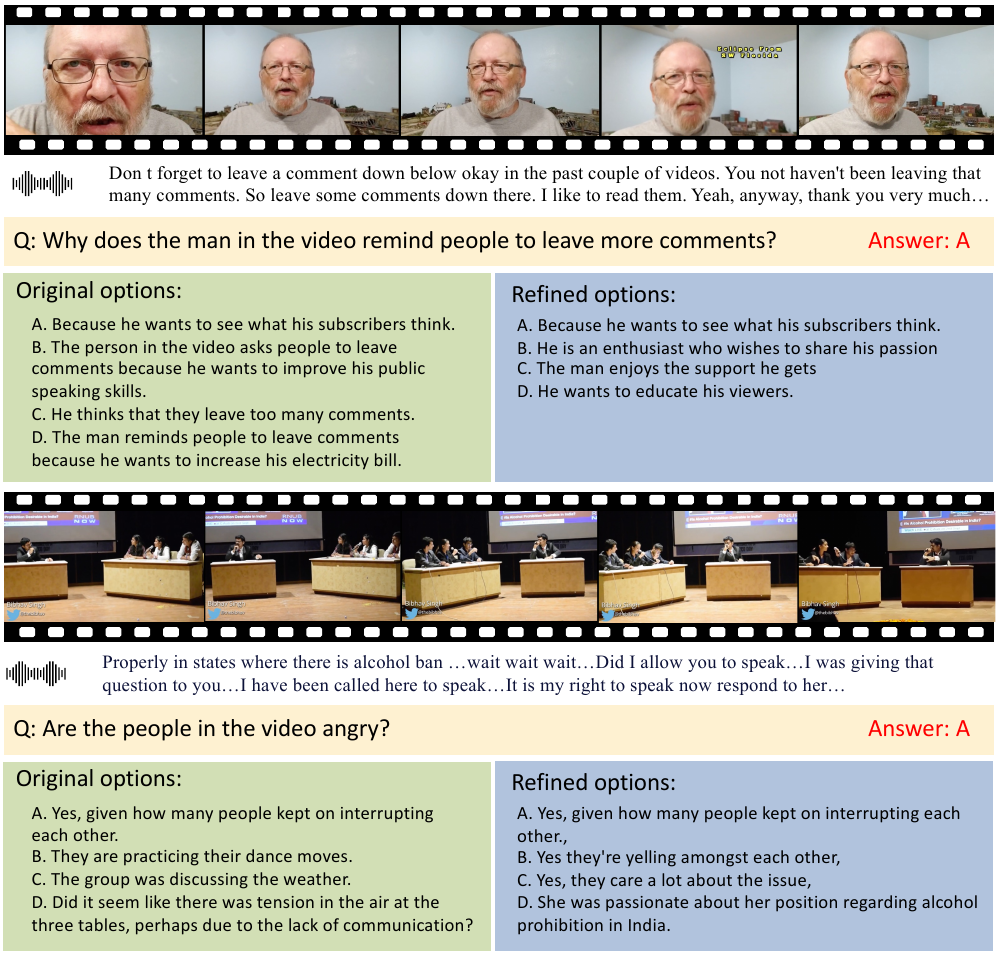}
    \caption{Comparison of the original options in Social-IQ 2.0 and our refined options.}
    \label{fig:comp}
\end{figure}

\begin{figure}[t!]
    \centering
    \includegraphics[width=0.8\linewidth]{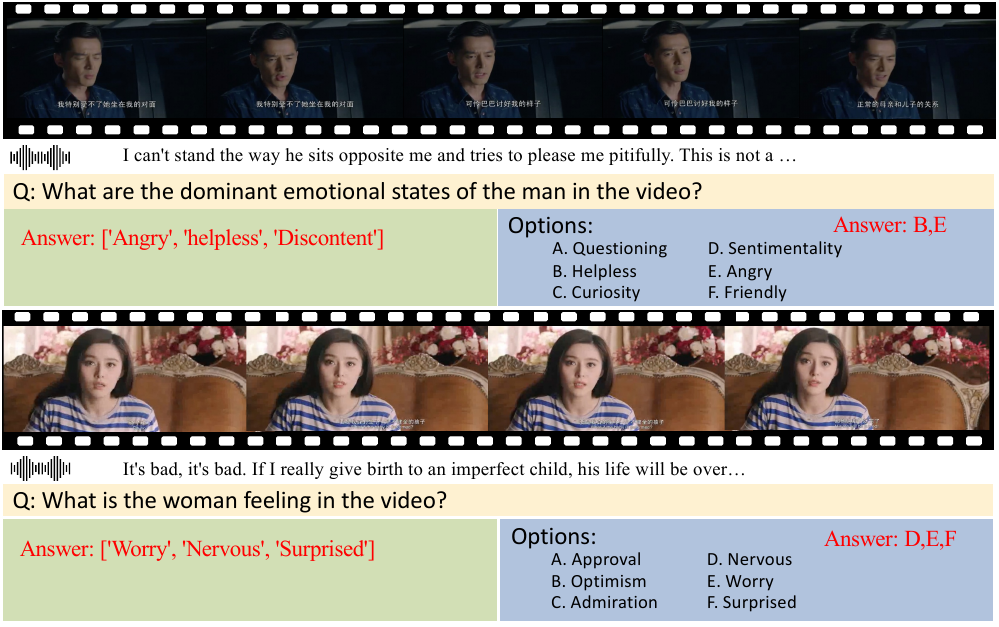}
    \caption{Comparison of the original open-vocabulary answer and our designed multi-choice question with multiple answers in EMER.}
    \label{fig:emer}
\end{figure}

In Social-IQ 2.0~\cite{siq2}, which is developed based on Social IQ 1.0\cite{zadeh2019social}, some questions in the dataset are straightforward and can be answered directly using only the text. Additionally, we also notice that some models may have been trained on this dataset.
Thus, we use GPT-4o~\cite{gpt4o} with only text modal for testing to identify challenging questions. We also replace easy options to increase the difficulty of the testing set. Finally, we conduct manual verification to ensure each question is relevant to multimodal information and cannot be answered directly through text alone. GPT-4o achieves 75.71\% on the original testing set. After our modifications, the performance is 60.02\%. 
Through these enhancements, the dataset becomes more challenging compared to its original version. Figure~\ref{fig:dataset} shows some comparisons between the original options and our refined version.


The original EMER~\cite{lian2023explainable} is not designed for open vocabulary evaluation. In the AffectGPT~\cite{lian2024affectgpt}, the authors propose grouping the predicted results and ground truth emotions for each sample using a large language model (LLM) and then calculating the F1 score. However, the need to frequently utilize LLMs for evaluation can lead to instability, causing significant challenges in both training and testing.
Thus, we refine the emotion vocabulary within these videos and organize all descriptive vocabulary into hierarchical categories. For the open vocabulary emotion options of the person in the videos, in addition to the original emotion ground-truth labels, we randomly select emotion description terms from other groups to create multiple-choice questions with multiple answers. Figure~\ref{fig:emer} shows some comparisons between the original answer and our refined version.

The MDPE dataset~\cite{cai2024mdpe} is initially crafted for classification tasks. To streamline the evaluation of multimodal models, we reformat them into a question-answer format and incorporate the original questions in each video as part of the query. Additionally, we refine the sample selection process by randomly selecting samples according to specific ratios based on difficulty levels.
We select samples where the interviewees feel uncertain about successfully deceiving—a total of 60 clips (i.e., the confidence rating is above 3). We also include 20 clips where they feel confident in their deception (i.e., the confidence rating is lower than 3, more challenging) and 120 no-deception clips, totaling 200 videos to form the deception data.

{
    \small
    \bibliographystyle{IEEEtran}
    \bibliography{main}
}

\end{document}